%% file: main.tex
\documentclass[runningheads]{llncs}
\usepackage{graphicx}
\usepackage{comment}
\usepackage{amsmath,amssymb} % define this before the line numbering.
\usepackage{color}

\usepackage[normalem]{ulem}
\usepackage{xcolor}

\DeclareMathOperator*{\argmax}{arg\,max}
\usepackage[ruled,vlined,linesnumbered,noresetcount]{algorithm2e}
\usepackage{appendix}
\newcommand{\resultval}{32.4\% }
\newcommand{\resulttest}{27.5\% }

\begin{document}
% \linenumbers
\pagestyle{headings}
\mainmatter
\def\ECCVSubNumber{3693}  % Insert your submission number here

\title{Deep Affinity Net: \\ Instance Segmentation via Affinity} 
% Replace with your title

\begin{comment}
    \titlerunning{ECCV-20 submission ID \ECCVSubNumber} 
    \authorrunning{ECCV-20 submission ID \ECCVSubNumber} 
    \author{Anonymous ECCV submission}
    \institute{Paper ID \ECCVSubNumber}
\end{comment}
%******************

% \begin{comment}
    \titlerunning{Deep Affinity Net}
    \author{
        Xingqian Xu\inst{1} 
        \and
        Mang Tik Chiu\inst{1}
        \and
        Thomas S. Huang\inst{1}
        \and
        Honghui Shi\inst{2,1}
    }
    \authorrunning{X. Xu, M. Chiu, T.S. Huang, H. Shi}
    \institute{
        {\small University of Illinois at Urbana-Champaign \and University of Oregon}
    }
% \end{comment}
%******************
\maketitle

\begin{abstract}

\input{TEX/abs.tex}


\end{abstract}

\input{TEX/body.tex}

\clearpage

\appendix
\input{TEX/appendix.tex}

\bibliographystyle{splncs04}
\bibliography{egbib}

\end{document}

%% file: TEX/abs.tex
%Most of the modern instance segmentation approaches fall into two categories: one is the region-based approach in which object bounding boxes are detected first and are later used in cropping and segmenting instances. The other is the keypoint-based approach in which individual instances are represented by a set of keypoints followed by a dense pixel clustering around those keypoints. Despite the maturity of these two paradigms, we would like to report an alternative approach where instances are segmented through dense affinity predictions followed by a graph partitioning algorithm. Such \textbf{affinity-based} approach foretells a potential resurrection on the paradigm where high-level graph features other than keypoints or regions can be directly applied in our segmentation task. Though it's accuracy lags behind the state-of-the-art methods, our end-to-end model results in AP \input{results/val.tex}on Cityscapes val and AP \input{results/test.tex}on test. Without bells and whistles, it achieves the best single-shot result among all affinity-based models.

Most of the modern instance segmentation approaches fall into two categories: region-based approaches in which object bounding boxes are detected first and later used in cropping and segmenting instances; and keypoint-based approaches in which individual instances are represented by a set of keypoints followed by a dense pixel clustering around those keypoints. 
Despite the maturity of these two paradigms, we would like to report an alternative \textbf{affinity-based paradigm} where instances are segmented based on densely predicted affinities and graph partitioning algorithms. 
Such affinity-based approaches indicate that high-level graph features other than regions or keypoints can be directly applied in the instance segmentation task.
In this work, we propose \textbf{Deep Affinity Net}, an effective affinity-based approach accompanied with a new graph partitioning algorithm \textbf{Cascade-GAEC}. 
Without bells and whistles, our end-to-end model results in \resultval AP on Cityscapes val and \resulttest AP on test. It achieves the best single-shot result as well as the fastest running time among all affinity-based models. It also outperforms the region-based method Mask R-CNN. 

%% file: TEX/body.tex
\section{Introduction}
\input{TEX/1_intro.tex}

%------------------------------------------------------------------------
\section{Related Work}
\input{TEX/2_related.tex}
%------------------------------------------------------------------------
\section{Our approach}
\input{TEX/3_method.tex}
%------------------------------------------------------------------------
\section{Experiments}
\input{TEX/4_experiments.tex}
%------------------------------------------------------------------------
\vspace{-0.1cm}
\section{Discussions}
\input{TEX/5_discussion.tex}
%------------------------------------------------------------------------
\vspace{-0.1cm}
\section{Conclusion}
\input{TEX/6_conclusion.tex}
%------------------------------------------------------------------------

%% file: TEX/1_intro.tex
Pixel affinity and graph partitioning algorithms are prevalent solutions towards image scene parsing tasks before the deep-learning era \cite{normalized_cut,szeliski2008comparative,crf_seg}. It was not until recent years that deep learning models promoted by DCNN \cite{dcnn} took over the spotlight. With the power of DCNN, generating image features that are embedded with rich graphic information is not only feasible but fast and convenient. As a result, many researchers are now concentrating on network design instead of algorithm design. However, in recent years' detection and segmentation works, there is a research trend of combining DCNN models with task-driven post-processing algorithms \cite{pfn,instance_cut,watershed,weekly_seed,weekly_rw,gmis,ssap}. For instance, Instance Cut \cite{instance_cut} and DWT \cite{watershed} use DCNN to identify instance edges and apply traditional algorithms for segmentation. GMIS \cite{gmis} utilizes both region proposals and pixel affinities to segment images. SSAP \cite{ssap} outputs the affinity pyramid and performs cascaded graph partition with some existing algorithms, which is similar to our approach.

\input{figures/cascade_gaec}

Region-based methods \cite{mask_rcnn,bshapenet,panet,upsnet}, also known as proposal-based methods, are descendants of detection networks such as \cite{spp,rcnn,fast_rcnn,faster_rcnn,fpn}. It is a prevalent and effective way to solve the instance segmentation task as an additional task from detection. However, these methods also share the same disadvantages such as multi-stage execution and high hyperparameter sensitivity. On the other hand, proposal-free methods \cite{pfn,instance_cut,watershed,josecb,deeperlab,ssap} are single-stage models that do not rely on bounding boxes. Some state-of-the-art models \cite{josecb,deeperlab} predict keypoints and use them as a strong hint for pixel-to-instance clustering. Based on the nature of these methods, we categorize them as keypoint-based methods because they use keypoints to tackle the instance segmentation problem. 

In addition to these two popular paradigms, we introduce Deep Affinity Net (DaffNet), and from that, we would like to promote another paradigm, namely affinity-based methods, to tackle instance segmentation from a different angle. In affinity-based methods, models will first predict affinities between different parts of the image, then parse the image through the graph partitioning algorithms. Previous works such as Instance Cut \cite{instance_cut}, GMIS \cite{gmis} and SSAP \cite{ssap} can be labeled as affinities-based methods because they share this similar design principle. Among them, the recent work SSAP \cite{ssap} reached a new state-of-the-art using multi-scale flipping inference, which demonstrated the effectiveness of this paradigm along with its potential. 

Our contribution in this work is to bring out heuristic solutions to conquer two major challenges in recent affinity-based methods: to predict affinities beyond fixed grid spacing and to reduce the running time of the NP-hard multicut graph partitioning problem \cite{corr_clutsering,corr_clutsering2}. With our newly designed grouping mechanism in DaffNet, we expand the network capacity in which affinities with arbitrary grid spacing can be inferred and utilized during the reconstruction of segments with disjoint parts. On the other hand, we leverage an existing greedy multicut algorithm, namely Greedy Additive Edge Contraction (GAEC) \cite{gaec}, into a modularized Cascade-GAEC with a theoretical complexity of $O(n^2\log n)$ (where $n$ is the number of pixels in the image). Moreover, with its cascaded structure, large object instances can be quickly allocated in low-resolution feature maps while small object instances can be segmented from higher resolution feature maps in later steps (Fig.~\ref{fig:cascade_gaec}). In the following chapter, we also provide evidence that the practical running time of Cascade-GAEC is approximately linear to the image size. In summary, the two highlights of our work are as follows: 

\begin{itemize}
    \item[$\bullet$] We introduce a novel grouping module for affinity-based methods in which affinities between any grid spacing can be inferred and utilized. Our unified affinity-based model, \textbf{DaffNet}, reaches \resultval AP on Cityscapes val and \resulttest AP on test with \textit{single-shot} inference method, which surpass all previous affinity-based methods. 
    \item[$\bullet$] We extend the greedy-based multicut algorithm GAEC \cite{gaec} into the newly designed \textbf{Cascade-GAEC}, which expedites graph partitioning on high-resolution images. The practical running time of Cascade-GAEC is approximately linear to image size and the absolute running time on one Cityscapes \cite{cityscapes} image is 0.24s, which is about $7\times$ faster than previous work \cite{ssap}.
\end{itemize}

%% file: figures/cascade_gaec.tex
\begin{figure}
    \vspace{-0.5cm}
    \centering
    \includegraphics[width=0.95\textwidth]{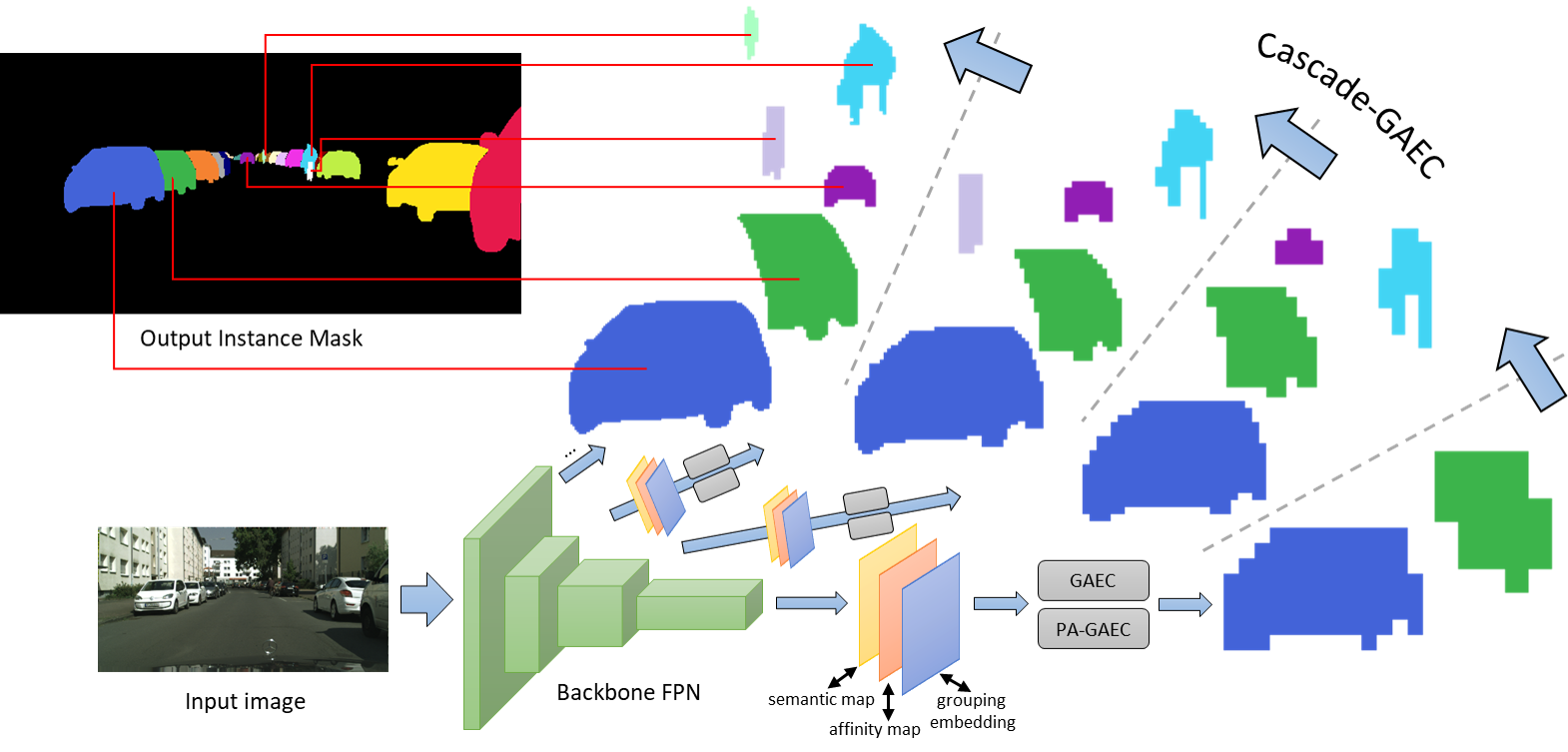}
    \vspace{-0.2cm}
    \caption{This graph shows the overall structure of our DaffNet and Cascade-GAEC. Given an input image, DaffNet utilizes its FPN backbone \cite{fpn} and computes semantic map, affinity map and grouping embedding with various resolutions. Then, it uses GAEC \cite{gaec} and Position-Aware GAEC (PA-GAEC) to cut out segments and pass them to the next level. Through this cascaded structure, large segments can be identified using low-resolution maps and its shape can be gradually refined. Small segments can be directly identified use high-resolution maps. }
    \vspace{-0.5cm}
\label{fig:cascade_gaec}
\end{figure}

%% file: TEX/2_related.tex
\label{sec:2}

\subsection{Instance Segmentation}

Previous works \cite{panet,upsnet} categorized region-based methods as proposal-based methods depending on whether a detector in models such as \cite{fast_rcnn,fpn,ssd,gfpn} is used to generate instance bounding boxes before segmenting instances. In earlier instance segmentation literature, SDS \cite{sds} used MCG \cite{mcg} for region proposals, CNN \cite{dcnn} for feature extraction and SVM for classification. MNC \cite{mnc} tackled the problem with three cascade network stages. Results such as bounding box, instance mask, and instance classification were orderly generated from one stage and fed to the next. FCIS \cite{fcis} predicted $k^2$ position-sensitive score maps where for each map, high scores were assigned to pixels with the same semantic class as well as the correct location of one proposal. As an extended version of Faster R-CNN \cite{faster_rcnn}, Mask R-CNN \cite{mask_rcnn} was a milestone two-stage instance segmentation network combining a region-proposal network \cite{faster_rcnn} with multi-purpose modules. Panoptic FPN \cite{panoptic_fpn}, extended from \cite{fpn}, leveraged the bottom-up pathway of the encoder network and effectively detected and parsed instances at the pyramid level that best-fitted the instance size.

Keypoint-based models, such as \cite{pfn,josecb,deeperlab}, first predict instance keypoints and later use them to group pixels by computing distances or using other clustering algorithms. For instance, in the training stage, PFN \cite{pfn} performs regression on all pixel embeddings and the normalized distances towards their instance bounding box corners. In the inference stage, pixels are grouped into instances using spectral clustering. Another work \cite{josecb} jointly optimizes the pixel spacial offset towards its instance center, the clustering margin on that center and the seed map for all potential instance centers. Recently, Deeperlab \cite{deeperlab} borrowed ideas from pose estimation models \cite{stacked_hourglass,person_lab} where long-short range offsets on pixel location towards both instance corners and centers are utilized for instance segmentation. 

Among affinity-based models \cite{instance_cut,gmis,ssap}, Instance Cut \cite{instance_cut} predicted instance edges and converted them into superpixels and affinities. It then applied a general multicut graph partitioning algorithm described in \cite{multicut,lmc} to segment images. In another work, GMIS \cite{gmis} generated instance segments using region proposals on top of the pixel affinities, making it both a region-based method and an affinity-based method. The recent work SSAP \cite{ssap} proposed a similar affinity pyramid structure in which affinities in a $5\times 5$ window are generated for each pixel and later are used to solve the optimization problem on graph partitioning. 

There exist other approaches towards instance segmentation that utilized pixel embedding \cite{deep_color}, watershed transform \cite{watershed} and recurrent neural nets \cite{crf_rnn,sgn,romera2016recurrent}. But these methods do not fall into the three main paradigms described above.

\subsection{Graph Partitioning Algorithm}
\label{sec:2_2}

The ideal approach towards solving the multicut problem is to generalize it into an integer programming problem. The formalized problem, namely Minimum Cost Multicut Problem \cite{nllmc,szeliski2008comparative}, can be defined as follows: Assume we have an undirected graph $G=(V,E)$ and a cost function $c: E\rightarrow \mathbb{R}$. Our best multicut solution is equivalent to finding the optimal zero-one solution for all $y_e\in Y_E$ in Eq.~\ref{eq:mcmp}, where $y_e=0$ means no-cut and $y_e=1$ means cut.

\begin{equation}\begin{gathered}
    \min_{y_e\in Y_E} \sum_{e\in E}c_e y_e \\ 
    \text{s.t.}\quad
    Y_E\in\{0,1\}^E \\
    \forall C \in \text{cycles}(G)\quad \forall e\in C: y_e\le \sum_{e'\in C\setminus \{e\}} y_e'
    \label{eq:mcmp}
\end{gathered}\end{equation}

When finding an optimal cut is not necessary, we can trade accuracy for efficiency using a greedy algorithm. Among which, Greedy Additive Edge Contraction (GAEC) \cite{gaec} is a fast and simple algorithm that better fits our graph partitioning requests on large images. The outline of GAEC is the following: The algorithm first assumes that all nodes are in the distinct partitions. It then performs greedy merge on two partitions that have the largest affinity value. After that, it updates all affinity values based on the merge. The algorithm terminates when no affinities are larger than the preset threshold. We also formulated GAEC in Alg.~\ref{alg:gaec}. Keep in mind that a larger affinity value $a_{u,v}$ means a higher chance that $u,v$ belongs to the same component, which is the reverse of $y_e$ in Eq.~\ref{eq:mcmp}.

Some other approaches such as spectral clustering \cite{sp_cluster}, normalized cut \cite{normalized_cut} and conditional random field (CRF) optimization \cite{crf_seg} share the similar graph partitioning principle in a high level. For instance, normalized cut solves graph partitioning through generalized eigenvalue decomposition with respect to the Laplacian matrices. CRF, on the other hand, are probabilistic models in which node labels embedded with graph partitioning information are inferred by optimizing the Gibbs energy function. However, these methods also sustain the long execution time when the size of the image increases.  
\vspace{-0.5cm}
\input{algorithms/gaec}
\vspace{-1cm}

%% file: algorithms/gaec.tex
\begin{algorithm}
    Set $G=(V,E)$\;
    Set $a_{u,v}$ = affinity of $e_{v,u} \in E$, $u,v\in V$\;
    Set $A$ = $\{a_{u,v}\}$\;
    $Q$ = Priority\_Queue($A$)\;
    \For {Q.top() $>$ threshold} {
        $a_e$ = $Q$.top()\; 
        $u_e, v_e$ = Vertex of $e$\;
        $u$ = merge($u_e, v_e$)\;
        Remove $e$ from $G$\;
        $V_{u_e},E_{u_e}$ = $V,E\in G$ connect with $u_e$\;
        $V_{v_e},E_{v_e}$ = $V,E\in G$ connect with $v_e$\;
        $V_{\cap}$ = $V_{u_e} \cap V_{v_e}$\;
        \For {v in $V_{\cap}$} {
            Add $e=\{u,v\}$, $a_{u,v}$=avg($a_{u_e,v}, a_{v_e,v}$) to $G$\;
            $Q$.insert($a_{u,v}$)\;
            $Q$.remove($a_{u_e,v}, a_{v_e,v}$)\;
        }
        Remove $E_{u_e},E_{v_e},u_e, v_e$ from $G$\;
        Add $u$ to $G$\;
    }
    \Return $G$\;
    \caption{GAEC}
    \label{alg:gaec}
\end{algorithm} 

%% file: TEX/3_method.tex
\label{sec:3}

We proposed a novel grouping mechanism for our affinity-based instance segmentation network, which is followed by an efficient greedy-based Cascade-GAEC algorithm for graph partitioning. Unified with the empirical FPN backbone \cite{fpn}, this new end-to-end model, named Deep Affinity Net(DaffNet), is capable to produce accurate results as well as to run faster compared with previous work \cite{ssap}. In this section, we will go through the design details of DaffNet and Cascade-GAEC. 
\input{figures/destructive_gaec_and_window}

%-----------------------------
\vspace{-0.1cm}
\subsection{Design Principle}
\label{sec:3_1}

The NP-hardness of finding an optimal multicut solution hinders researchers from parsing high-resolution images on large modern datasets through affinity-based approaches. Therefore, our fundamental design principle in DaffNet is to fuse the efficient greedy algorithm GAEC into our affinity-based approach without losing accuracy. 

\textbf{Low error tolerance} of GAEC on its input is the most challenging problem for DaffNet. This is because that the greedy algorithm makes optimal decisions locally instead of looking for a global optimum. Therefore, in GAEC and the Cascade-GAEC in this work, a wrongly predicted affinity could be destructive because two unrelated segments can be falsely merged (Fig.~\ref{fig:destructive_gaec_and_window}). Therefore, comparing with GMIS \cite{gmis} and SSAP \cite{ssap}, DaffNet limits its ability on predicting fixed grid spacing affinities to only the 4 nearby neighbors (Fig~.\ref{fig:destructive_gaec_and_window}). Such a design enables the network to focus only on predicting neighboring affinities, and thus increases its accuracy dramatically. 

\textbf{Disjointed instance segments}, however, cannot be regrouped based on neighboring affinities. Object instances are frequently segregated into disconnected pieces due to occlusions. Previous works \cite{gmis,ssap} partially solved this issue by predicting fixed-grid affinities between non-adjacent pixels. Here in DaffNet, we propose an extra grouping module where disjointed image partitions with arbitrary distances can be regrouped based on the similarity of their predicted embeddings. We are going to explain the details of the grouping module in the next subsection.

%-----------------------------
\vspace{-0.1cm}
\subsection{The Grouping Module}

As described earlier, the grouping module is designed to infer the affinity between non-connected segments with arbitrary grid spacing. At each pyramid level, the grouping module generates a $k\times h\times w$ feature map using two $3 \times 3$ convolution layers and one $1 \times 1$ convolution layer. Each of the two $3 \times 3$ convolution layers is followed by In-Place Activated BatchNorm (Inplace-ABN) \cite{inplace_abn} and leaky ReLU with a negative slope of 0.01. This $k\times h\times w$ feature map provides us a $k$-dimension embedding at each pixel location. Inspired by CornerNet \cite{cornernet}, we adopt the ``push-pull'' principle during the training phase. The goal is to ``push'' pixel embeddings away from each other if they belong to two different instances, and to ``pull'' pixel embeddings closer if they belong to the same instance. We first compute mean embeddings for all segments through averaging their pixel embeddings. We then minimize the difference between mean embedding and pixel embeddings from the same instance. Lastly, we maximize the difference between all pairs of mean embeddings from different instances. The entire process is also explained in Fig.~\ref{fig:grouping_module}. 

\input{figures/grouping_module}

\subsection{The Grouping Loss}

Different from CornerNet \cite{cornernet}, when we quantify the similarity between two embeddings $x_a$ and $x_b$, we do not regress on their $l^2$ distance. Instead, we compute the Gaussian function $\Phi_{a,b}$ between the two embeddings as shown in Eq.~\ref{eq:guassian_distance}. Through computing $\Phi_{a,b}$, we convert the $[0, +\infty)$ distance into a $(0, 1]$ score that measures the affinity between $a$ and $b$. The hyperparameter $\alpha$ is set to $\ln(2)$ so that when $||x_a-x_b||_2^2>1$, $\Phi_{a,b}<0.5$. 

\begin{equation}
    \Phi_{a,b} = \Phi(x_a, x_b) = \exp\left(-\alpha (x_a-x_b)^T(x_a-x_b)\right)
    \label{eq:guassian_distance} \in (0, 1]
\end{equation}

Combined with the ``push-pull'' principle we described earlier, we can then formulate the loss function in Eq.~\ref{eq:grp_loss}. $L_{gd}$ represents the push loss and $L_{gs}$ represents the pull loss. $x_i$ represents the embedding for pixel $i$. $S$ represents an instance segment and $N_S$ means the number of pixels in $S$. The mean embedding of $S$: $x_S = \frac{1}{N_{S}}\sum_{i\in S}x_i$. $N_{ins}$ is the total number of instances. 

\begin{equation}
    \begin{gathered}
    L_{gd} = \frac{1}{N^2_{ins}-N_{ins}}\sum_{S\ne S'}
        \text{BCE}\left(0, \Phi(x_S, x_{S'})\right)\\
    L_{gs} = \frac{1}{N_{ins}}\sum_{i=1}^{N_{ins}}
        \frac{1}{N_{S}}\sum_{i \in S}
        \text{BCE}\left(1, \Phi(x_S, x_i) \right)
    \end{gathered}
\label{eq:grp_loss}
\end{equation}

%-----------------------------
\vspace{-0.1cm}
\subsection{Deep Affinity Net}

Besides the grouping module, DaffNet also predicts a $c\times h \times w$ semantic map and a $4\times h\times w$ affinity map on each pyramid level using the semantic and affinity module. $c$ is the number of semantic categories and 4 means the 4 neighbor affinities. The semantic and affinity module also consist of two $3 \times 3$ convolution layers and one $1 \times 1$ convolution layer followed by Inplace-ABN and leaky ReLU. In the affinity module, to ensure that the affinities between two pixels are the same in both directions, we set both affinities to their average value.

During the training phase, we compute the cross-entropy loss $L_{sem}$ on the semantic map. For the affinity module, we separately compute two binary cross-entropy losses: $L_{ab}$ and $L_{as}$, which represent the losses of pixels that fall on the boundary of any object and those that do not, respectively. To formulate these losses, we set $N,N_{bdr},N_{sld}$ as the total pixel number, the boundary pixel number and the non-boundary pixel number, where $N_{bdr}+N_{sld}=N$. Let $p_{n,c},q_{n,c}$ be the predicted semantic score and ground truth on pixel $n$ with class $c$, and let $\hat{p}_{n,d}, \hat{q}_{n,d}$ be the predicted and ground truth affinities. The losses are computed with the following equations:

\begin{equation}
    \begin{gathered}
        L_{sem} = -\frac{1}{N}\sum_{n=1}^{N}\sum_{c=1}^{C}q_{n,c}\log(p_{n,c})\\
        L_{ab} = \frac{1}{N_{P_B}}\sum_{n=1}^{N_{P_B}}w_n\sum_{d=1}^{4}
            \text{BCE}(\hat{q}_{n,d}, \hat{p}_{n,d})\\
        L_{as} = \frac{1}{N_{P_S}}\sum_{n=1}^{N_{P_S}}w_n\sum_{d=1}^{4}
            \text{BCE}(\hat{q}_{n,d}, \hat{p}_{n,d})\\
    \end{gathered}
    \label{eq:sem_aff_loss}
\end{equation}

The final loss of DaffNet is then a weighted sum of $L_{sem}, L_{ab}, L_{as}, L_{gd}$ and $L_{gs}$ from all levels of the pyramid. 

%---------------------------------
\vspace{-0.1cm}
\subsection{Cascade-GAEC}
\label{sec:3_5}

Recall that our Cascade-GAEC is an extended version of the multicut algorithm GAEC \cite{gaec} (Sec.~\ref{sec:2_2}, Alg.\ref{alg:gaec}). The key idea of Cascade-GAEC is to parse instances from the low-resolution affinity map and then to gradually refine its shape from coarse to fine. Also, small instances that cannot be identified using the low-resolution affinity map can be captured using the higher resolution affinity map (Fig.~\ref{fig:cascade_gaec}). Thus, when comparing with GAEC \cite{gaec}, Cascade-GAEC can better handles object sizes because large and small instances are separately parsed in different pyramid levels. In addition, Cascade-GAEC runs much faster than regular GAEC. This is because the top-down information flow of Cascade-GAEC helps pre-grouping most parts of the image and thus reduces the workload when comparing with finding fine instance segments directly from high-resolution images. 

An intermediate step of the Cascade-GAEC is to upsample low-resolution partitioning results into high-resolution results, and then it serves as an input to the next level GAEC. Since the height and width of one pyramid level of the FPN is exactly half of the next pyramid level. The upsampling process expands the original labeling $4\times$ larger. Then, pixels that fall on objects boundaries are reclaimed as unlabeled pixels. This is shown in Fig.\ref{fig:upsampling}. 

\input{figures/upsampling}

The outputs of DaffNet include both neighboring affinities and affinities for arbitrary grid spacing. To accommodate this change, on each pyramid level, we perform the grouping algorithm twice. One is regular GAEC that helps to group pixels into segments, the other is a newly designed Position-aware GAEC (PA-GAEC) in which disconnected segments are merged based on the predicted semantic map and grouping embeddings. An important feature of PA-GAEC is that the input affinities are regulated according to the physical location of the segment. Segments far apart from each other will be weighed down, and thus have a lower probability to be merged. A more detailed illustration of PA-GAEC can be found in Alg.\ref{alg:pa_gaec} where the ``position-aware'' part is highlighted in red. 

\vspace{-0.5cm}
\input{algorithms/pa_gaec}
\vspace{-0.5cm}

To further accelerate Cascade-GAEC, we design a helper function called Greedy  Association (GAS). The idea on GAS is to avoid the maximum affinity search and to iteratively assign unlabeled pixels with the same label as its closest related neighbor until no more assignment can be done. GAS can substitute the bottom-level GAEC and reduce the burden of computing the finest instance segments. The complete GAS algorithm is in Appendix~\ref{appex:1}. 

% \vspace{-0.5cm}
% \input{algorithms/gas}
% \vspace{-0.5cm}

Finally, the full algorithm of Cascade-GAEC is shown in Alg.\ref{alg:cascade_gaec}. In our experiments, $\beta$ in PA-GAEC is set to 0.5, all thresholds in the algorithm are set to 0.5.

\vspace{-0.5cm}
\input{algorithms/cascade_gaec}
\vspace{-1cm}

%% file: figures/destructive_gaec_and_window.tex
\begin{figure}
    \centering
    \includegraphics[width=0.8\textwidth]{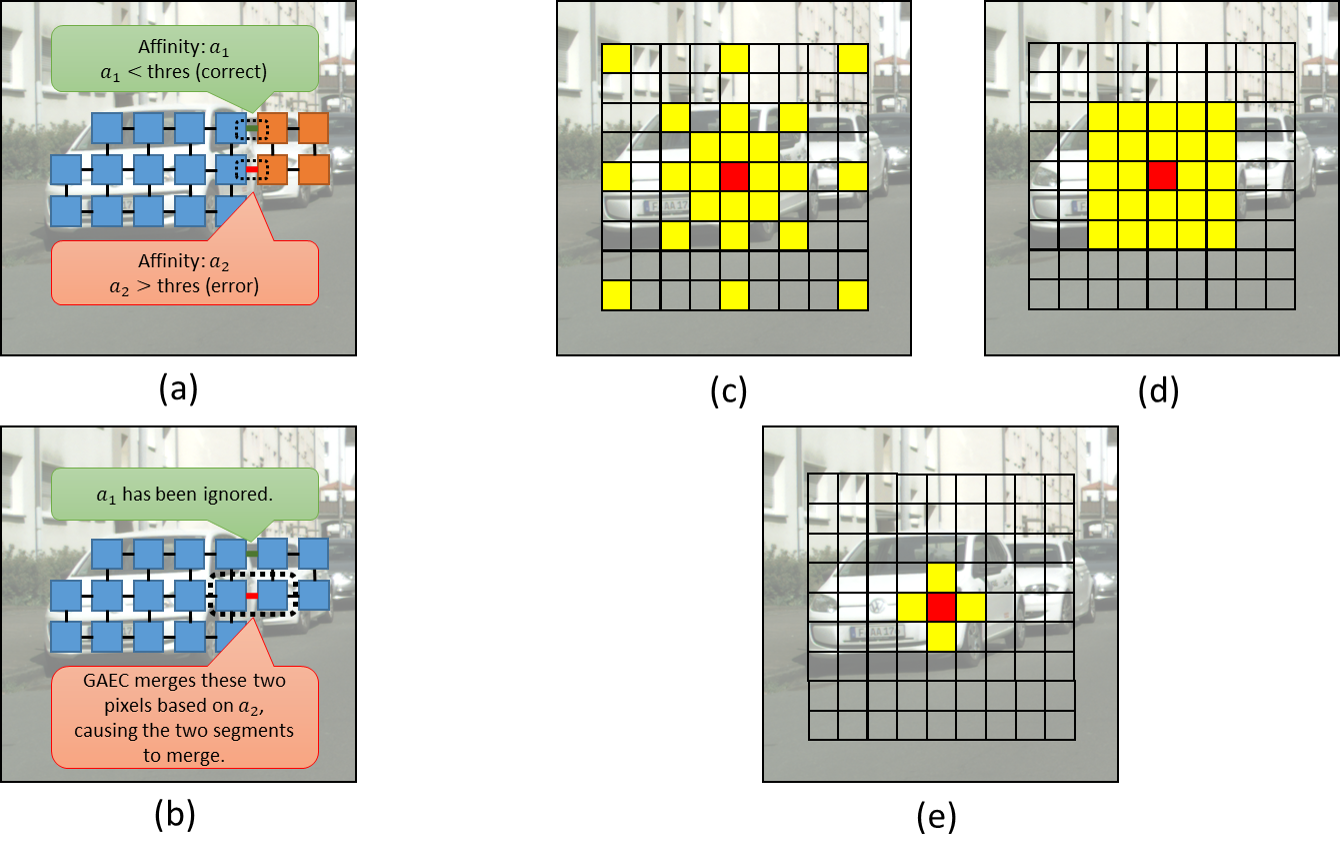}
    \vspace{-0.2cm}
    \caption{Figure (a) and (b) show an fail case using GAEC in which one erroneous affinity prediction destroys two segments by falsely merging them together. Figure (c), (d) and (e) show the fixed grid affinities predicted by GMIS \cite{gmis}, SSAP \cite{ssap} and DaffNet (ours) respectively. Red boxes are the center pixels and yellow boxes are the target pixels. }
    \vspace{-0.5cm}
\label{fig:destructive_gaec_and_window}
\end{figure}

%% file: figures/grouping_module.tex
\begin{figure}
\vspace{-0.5cm}
\centering
    \includegraphics[width=0.95\textwidth]{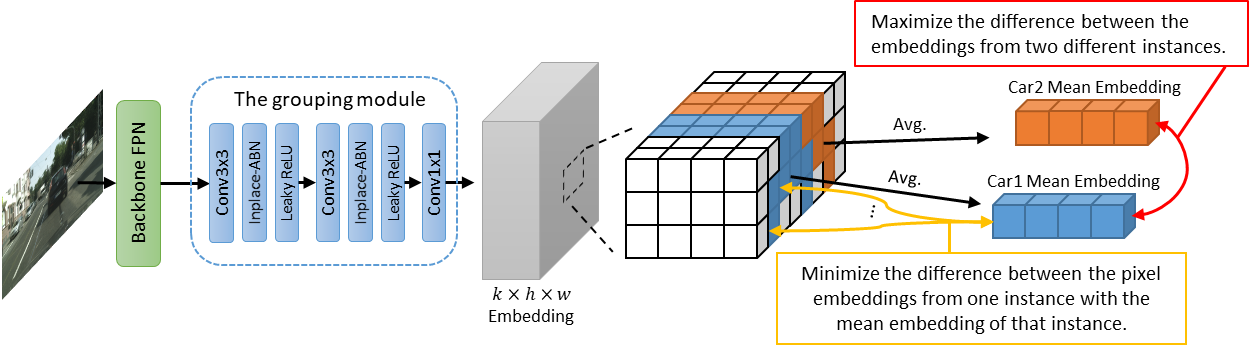}
    \vspace{-0.2cm}
    \caption{This figure shows the structure of our grouping module. It also illustrates the general ``push-pull'' principle that we use to generate the loss functions on the predicted embeddings. }
    \vspace{-1cm}
\label{fig:grouping_module}
\end{figure}

%% file: figures/upsampling.tex
\begin{figure}
    \centering
    \includegraphics[width=0.85\textwidth]{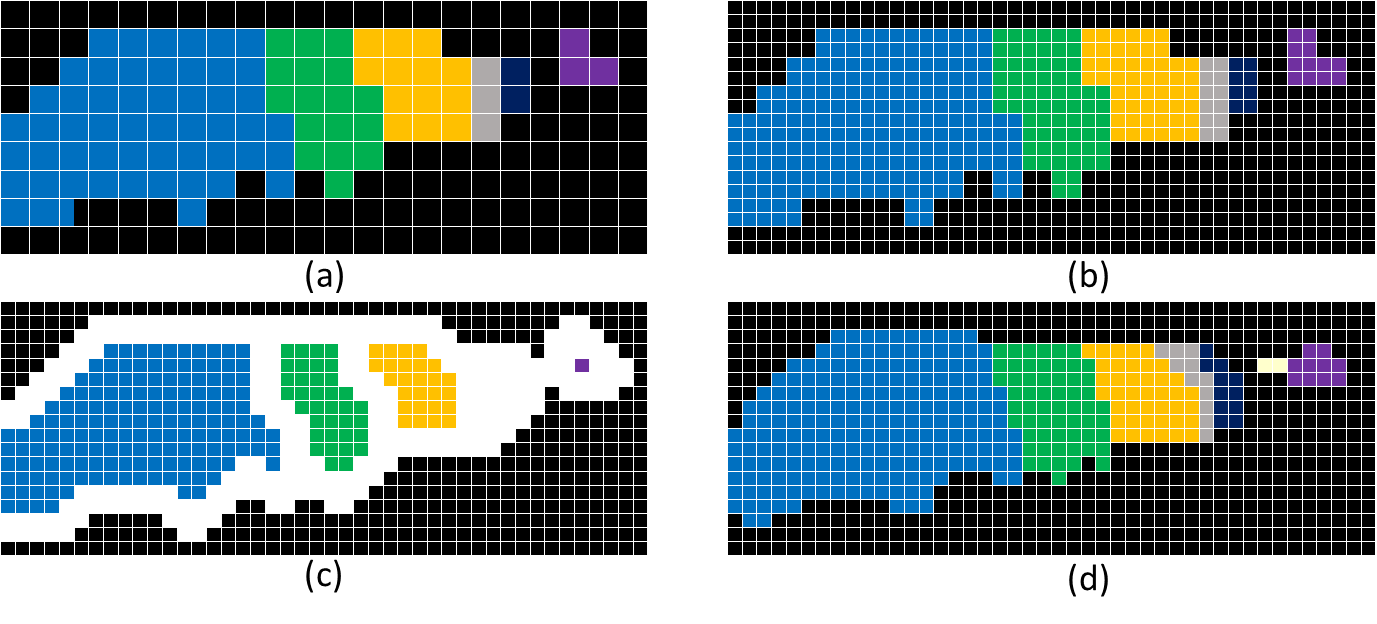}
    \vspace{-0.5cm}
    \caption{This figure explains how low-resolution segments are upsampled into high-resolution segments and are delivered to the next pyramid level in Cascade-GAEC. (a) is the segmentation map at some pyramid level. Black pixels are background and colored pixels are segments. (b) is directly created from a nearest upsampling on (a). The height and width in (b) are twice as (a). In (c), pixels are unlabelled (white) if it is on the boundary of segments. (c) then serves as an input of GAEC at the next pyramid level. (d) shows the refined result after GAEC.}
    \vspace{-0.5cm}
\label{fig:upsampling}
\end{figure}

%% file: algorithms/pa_gaec.tex
\begin{algorithm}
    Set $G=(V,E)$\;
    Set $a_{u,v}$ = affinity of $e_{v,u} \in E$, $u,v\in V$\;
    \textcolor{red}{Let $h_i, w_i, ch_i, cw_i$ = height, width, center y and x of vertex $i, \forall i\in V$}\;
    \textcolor{red}{Let $d_{u,v}$ = $
        \min\left(1, \frac{0.5\times\max(h_u,h_v)}{|ch_u-ch_v|}\right)^{\beta}
        \times
        \min\left(1, \frac{0.5\times\max(w_u,w_v)}{|cw_u-cw_v|}\right)^{\beta}
    $}\;
    \textcolor{red}{Set $A, D$ = $\{a_{u,v}\}, \{d_{u,v}\}$}\;
    $Q$ = Priority\_Queue(\textcolor{red}{$A\odot D$})\;
    \For {Q.top() $>$ threshold} {
        GAEC\_loop()\;
        \textcolor{red}{Update($A_{new},D_{new}$)}\;
        \textcolor{red}{$Q$.insert($A_{new}\odot D_{new}$)}\;
    }
    \Return $G$\;
    \caption{Position-aware GAEC (PA-GAEC)}
    \label{alg:pa_gaec}
\end{algorithm} 

%% file: algorithms/cascade_gaec.tex
\begin{algorithm}
\caption{Cascade-GAEC}
    \For {$i$ = $N_{plvl}$ to 1} {
        \If {$i \ne N_{plvl}$} {
            $G^{(i)}$ = upsample($G^{(i+1)}$)\;
            $A_a \leftarrow \hat{p}^{(i)}$, $\hat{p}^{(i)}$ is the affinity map\;
            \If {$i=1$} {
                $G^{(i)}$ = GAS($G^{(i)}, A_a$)\;
                \Return $G^{(i)}$\;
            }
        }
        $G^{(i)}$ = GAEC($G^{(i)}, A_a$)\;
        $A_s \leftarrow $ Jensen\_Shannon\_Divergence($p^{(i)}$), $p^{(i)}$ is the semantic map\;
        $A_g \leftarrow \Phi(x^{(i)})$, $x^{(i)}$ is the embedding map\;
        $G^{(i)}$ = PA\_GAEC($G^{(i)}, A_s\odot A_g$)\;
    } 
\label{alg:cascade_gaec}
\end{algorithm} 

%% file: TEX/4_experiments.tex
%%%%%%%%%%%%%%%%%%%%%%%
% Dataset and Metrics %
%%%%%%%%%%%%%%%%%%%%%%%

\subsection {Dataset and Metric}

We use Cityscapes \cite{cityscapes} as our training and evaluation dataset. Cityscapes contains 25000 $1024\times2048$ urban street scene images labeled on a total of 19 semantic classes, including 8 instance classes. The 25000 images are further split into 5000 finely annotated images and 20000 coarsely annotated images. In our experiment, we only use the finely annotated images for training. Like in other instance segmentation works, our evaluation metric is the Average Precision (AP) which is computed by averaging the precision under Intersection over Union (IoU) thresholds from 0.5 to 0.95 with a 0.05 step increment. 

%%%%%%%%%%%%%%%%%%%
% Network Details %
%%%%%%%%%%%%%%%%%%%%

\vspace{-0.1cm}
\subsection {Implementation Details}

The encoder structure of the FPN \cite{fpn} backbone in DaffNet is ResNet-101 \cite{resnet}. We keep most of the ResNet structure unchanged so we end up having 4 feature maps with $\frac{1}{4^2}, \frac{1}{8^2}, \frac{1}{16^2}, \frac{1}{32^2}$ of the original input size and $256, 512, 1024, 2048$ channel numbers. These feature maps are the immediate outputs after the ResNet layer block 1 to 4. The only modification in ResNet is that we use dilation rate 4 in the last layer block (layer4). This dilated layer block has the same structure as the layer block in the Deeplab series \cite{deeplabv1,deeplabv2,deeplabv3,deeplabv3p}. Besides, our FPN backbone has customized lateral connections and upsampling paths. The lateral structure is composed of two $3\times 3$ conv layer and one $1\times 1$ conv layer with two Inplace-ABNs \cite{inplace_abn} and leaky ReLUs in between. We use negative slop $\alpha=0.01$ on leaky ReLU for the entire network. These lateral layers maintain the channel number and they only exist in the bottom three pyramid levels. The upsampling path is composed of one deconvolution layer, one Inplace-ABN, one leaky ReLU, and one conv1x1 layer. The upsampling path and the lateral path are merged together through addition. After that, the feature map passes through another two $3\times 3$ conv layer, Inplace-ABN and leaky ReLU to produce the shared output for other modules. The structure of semantic, affinity and grouping modules are mentioned in Sec.~\ref{sec:3}. 

%%%%%%%%%%%%
% Training %
%%%%%%%%%%%%

\vspace{-0.1cm}
\subsection {Training}

There are two phases in our training process. In the first phase, we use the ImageNet pretrained model and train a similar network on Cityscapes with only semantic loss. We reach 76.76\% mIoU on the semantic task on the Cityscapes val set. In the second phase, we use the phase-one model as the phase-two pretrained model. Our data augmentation process can be described as the follows: We randomly scale the image by a factor uniformly drawn from $[0.7, 1.3]$; We then random crop a $513\times1025$ window followed by random flipping; Lastly, we normalize the image using its color mean and variance. The following loss weights are used in training ($\lambda$ are the corresponding weights on the losses. For example, $\lambda_{sem}^{(1)}$ is the weights on $L_{sem}$ on the first pyramid level.): 

\begin{enumerate}
    \itemsep0em
    \item Semantic: $\lambda_{sem}^{(i)}$=2 for any pyramid level $i$;
    \item Grouping: $\lambda_{gd}^{(i)}$=$\lambda_{gs}^{(i)}$=0.5 for any pyramid level $i$;
    \item Affinity: $\lambda_{ab}^{(4)}$=$\lambda_{as}^{(4)}$=$\lambda_{as}^{(3)}$=$\lambda_{as}^{(3)}$=1;
    \item Affinity: $\lambda_{ab}^{(2)}$=$\lambda_{as}^{(2)}$=0.5, $\lambda_{ab}^{(1)}$=$\lambda_{as}^{(1)}$=0.25;
\end{enumerate}

We run our experiments with a batch size of 16 on 4 RTX 2080Ti GPUs for 55500 iterations (300 epochs). We set our base learning rate to 0.01 and weight decay to 5e-4 for the SGD optimizer. We use 1000 iterations linear warm-up \cite{linear_warmup} followed by another 10000 iterations constant learning rate at 0.01. We then use the poly rule \cite{deeplabv3p} to slowly decrease the learning rate from 0.01 to 0 in the last 44500 iterations. 

%%%%%%%%%%%%%%%%%%%%%
% Experiment Result %
%%%%%%%%%%%%%%%%%%%%%

\input{tables/final.tex}

\vspace{-0.1cm}
\subsection {Quantitative and Qualitative Results}

Using the hyperparameters and the training pipeline we described in the last subsection, we trained a DaffNet model that reaches \resultval AP on Cityscapes val and \resulttest AP on test. During the evaluation, we applied Cacade-GAEC without GAS. No further post-processing techniques such as flip, multi-scale and model ensemble were applied. This setting aligns with our design principles in which we target on exploring efficient and accurate affinity-based methods. (Section \ref{sec:3_1}) 

We also notice that DaffNet has the best performance on large object instances such as bus, truck, and train. Such behavior could be quite critical for real-world applications because large and nearby traffic participants are what we need to notice first for safety. A more detailed performance comparison between our model and other instance segmentation models is listed in Table \ref{table:final}. Some of our segmentation results are visualized in Figure \ref{fig:demo}.

\input{figures/demo}

%%%%%%%%%%%%%%%%%%%%%%%%
% Algorithm Efficiency %
%%%%%%%%%%%%%%%%%%%%%%%%

\subsection {Algorithm Efficiency Analysis}

Cascade-GAEC is an efficient algorithm mainly because it is a greedy algorithm. Recall that the worst-case complexity of GAEC is $O(n^2\log n)$ where $n$ is the number of vertices \cite{gaec}. Our modified PA-GAEC has exactly the same complexity because finding the new positions of the merged segments is $O(1)$ complexity. Cascade-GAEC is a combination of a finite number of GAEC and PA-GAEC, thus it inherits the theoretical worst-case complexity of $O(n^2\log n)$. In DaffNet, Cascade-GAEC works with only four affinities per pixel, and with far less pairwise affinities from the grouped segments. Thus, the actual complexity of Cascade-GAEC in DaffNet is about $O(n\log n)$. In practice, Cascade-GAEC is further expedited due to its cascade structure. The real-world running time has an approximately linear relationship with the number of pixels in the input image and takes only 0.244s on a $1024\times 2048$ Cityscapes image (Figure~\ref{fig:running_time}). In the latest affinity-based work SSAP \cite{ssap}, the fastest running time is approximately 0.5s per $6\times10^5$ pixels, equivalent to roughly 1.74s per Cityscapes image according to their chart. Thus our Cascade-GAEC is approximately 7 times faster than SSAP.

\input{figures/running_time}

%%%%%%%%%%%%%%%%%%%%%%%%%%%%%%%%%%%%%%%%
% Hyperparameters and Ablation Studies %
%%%%%%%%%%%%%%%%%%%%%%%%%%%%%%%%%%%%%%%%

\vspace{-0.1cm}
\subsection {Hyperparameter and Ablation Study}

In Table~\ref{table:experiment_a} and \ref{table:experiment_b}, we evaluate the performance of DaffNet using different dimension $k$ of grouping embedding and different weights for the affinity BCE loss (Eq.~\ref{eq:sem_aff_loss}). As shown in the table, we try three $k$s: 8, 16 and 32, and 4 combinations of weights. Our conclusion on $k$ is that the higher the dimension we use the better the AP results are. Moreover, decaying weights: $0.25,0.5,1,1$ on affinity losses gives us the best result. 

\input{tables/experiment_a_b}

Our next experiment is an ablation study on our Cascade-GAEC algorithm. The experiment is carried out by evaluating the same model with different GAEC setups (Table~\ref{table:experiment_c}). We first generate a baseline AP 26.6\% without using the grouping modules. Then, we add back the grouping module and calculate its results via GAEC. This provides us a 4.2\% AP increment. Next, we use our PA-GAEC instead of GAEC and we increase the AP by an additional 1.6\%. Finally, we remove the speed-boosting GAS and compute the finest solution using the regular GAEC and PA-GAEC combination. This gives us an additional 0.4\% AP increment.

\input{tables/experiment_c}

%% file: tables/final.tex
\begin{table}[t!]
\centering
\caption{This table compares our results with other models in the Cityscapes instance segmentation challenges. For previous works, we put them into three categories: region-based, keypoint-based and affinity-based. }
\resizebox{0.98\textwidth}{!}{
    \begin{tabular}{l|c|cccc|cccccccc}
    \hline
    Method                            & AP(Val) & AP(Test) & AP50\%  & AP50m   & AP100m  & Person & Rider & Car  & Truck & Bus  & Train & Motorcycle & Bicycle\\ \hline\hline
    
    \textit{region-based}             &         &          &         &         &         &        &       &      &       &      &       &            &        \\ 
    Mask R-CNN \cite{mask_rcnn}       & 31.5    & 26.2     & 49.9    & 40.1    & 37.6    & 30.5   & 23.7  & 46.9 & 22.8  & 32.2 & 18.6  & 19.1       & 16.0   \\ 
    GMIS \cite{gmis}                  & 34.1    & 27.6     & 44.6    & 47.9    & 42.7    & 31.5   & 25.2  & 42.3 & 21.8  & 37.2 & 28.9  & 18.8       & 12.8   \\ 
    PANet (fine only) \cite{panet}    & 36.4    & 31.8     & 57.1    & 46.0    & 44.2    & 36.8   & 30.4  & 54.8 & 27.0  & 36.3 & 25.5  & 22.6       & 20.8   \\ 
    UPSNet \cite{upsnet}              & -       & 33.0     & 59.6    & 50.7    & 46.8    & -      & -     & -    & -     & -    & -     & -          & -      \\ \hline\hline
    
    \textit{keypoint-based}           &         &          &         &         &         &        &       &      &       &      &       &            &        \\ 
    Multi-task \cite{multi_task}      & -       & 21.6     & 39.0    & 37.0    & 35.0    & 19.2   & 21.4  & 36.6 & 18.8  & 26.8 & 15.9  & 19.4       & 14.5   \\ 
    Jointly-Optim \cite{josecb}       & -       & 27.6     & 50.9    & 37.3    & 37.8      & 34.5   & 26.1  & 52.4 & 21.7  & 31.2 & 16.4  & 20.1       & 18.9   \\ \hline\hline
    
    \textit{affinity-based}           &         &          &         &         &         &        &       &      &       &      &       &            &        \\ 
    Instance cut \cite{instance_cut}  & -       & 13.0     & 27.9    & 26.1    & 22.1    & 10.0   & 8.0   & 23.7 & 14.0  & 19.5 & 15.2  & 9.3        & 4.7    \\ 
    GMIS (no RoIs \& Flip) \cite{gmis} & 22.8    & -        & -       & -       & -       & -      & -     & -    & -     & -    & -     & -          & -      \\ 
    SSAP (no MS \& FLIP) \cite{ssap}  & 31.5    & -        & -       & -       & -       & -      & -     & -    & -     & -    & -     & -          & -      \\ 
    DaffNet (ours with GAS)                    & 32.0    & 27.1     & 46.9    & 46.6    & 41.2    & 24.5   & 20.8  & 42.7 & 29.2  & 38.3 & 32.2  & 17.9       & 10.9   \\
    DaffNet (ours)              & \textbf{32.4}    & \textbf{27.5}     & 48.0    & 46.9    & 41.5    & 24.5   & 22.2  & 43.7 & 29.5  & 38.3 & 31.9  & 18.0       & 12.1   \\ \hline
    \end{tabular}
}
\label{table:final}
\vspace{-0.5cm}
\end{table}

%% file: figures/demo.tex
\begin{figure}
    \vspace{-0.5cm}
    \centering
    \includegraphics[width=\textwidth]{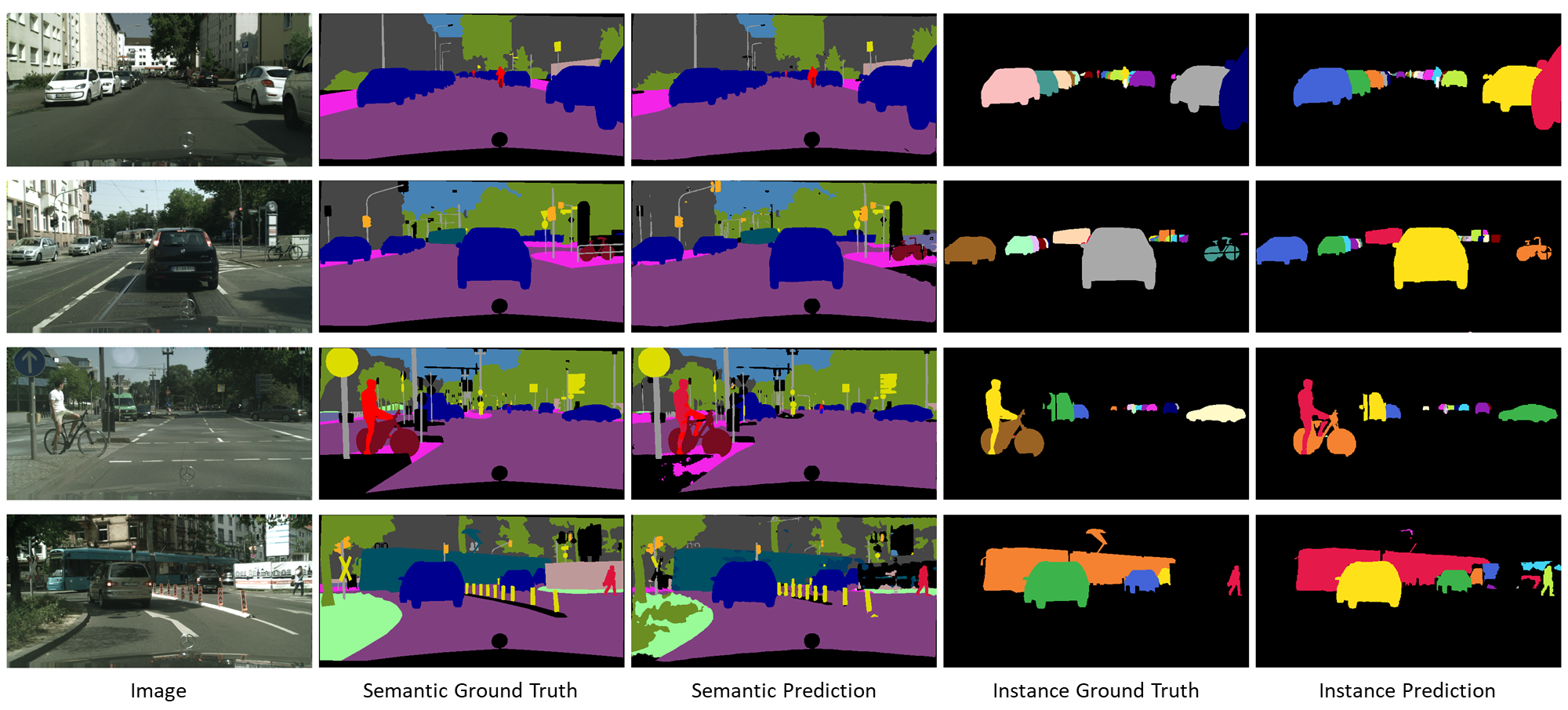}
    \vspace{-0.5cm}
    \caption{This figure shows the visualization of DaffNet's results. Our model is capable to accurately recover the fine contours of both nearby and distant instances from all categories. }
    \vspace{-1cm}
\label{fig:demo}
\end{figure}

%% file: figures/running_time.tex
\begin{figure}
    \centering
    \vspace{-0.5cm}
    \includegraphics[width=0.8\linewidth]{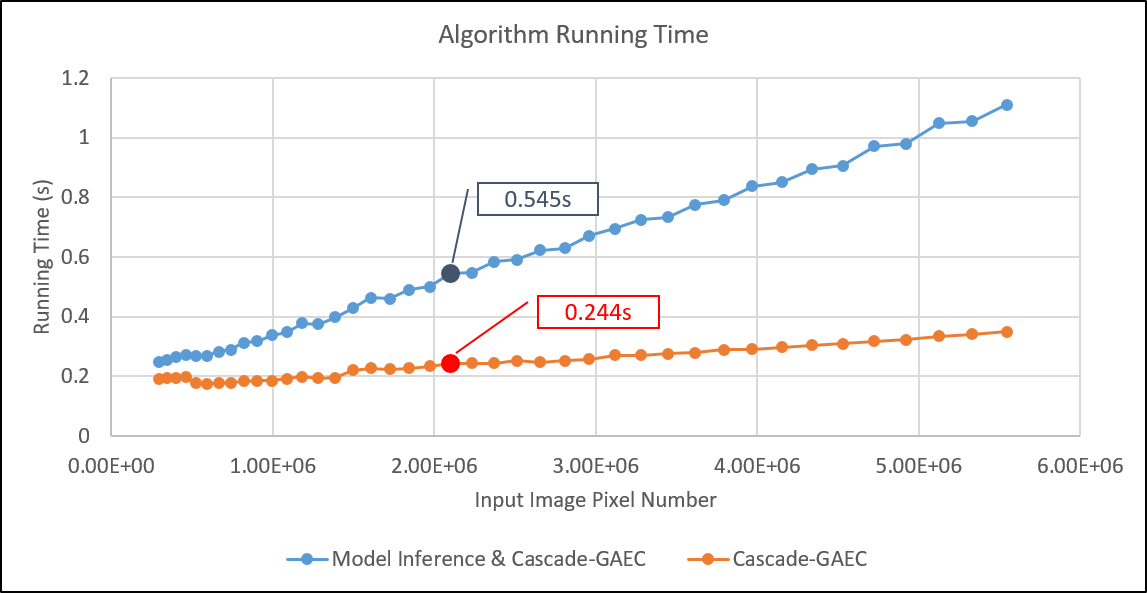}
    \caption{Cascade-GAEC running time on a regular desktop with one Intel-i7-8700K CPU and one GTX 1080 Ti GPU. The evaluate batchsize is 4 and the CPU thread number is also 4. The two bold dots with number boxes show the running time for a regular Cityscapes image. }
    \vspace{-1cm}
\label{fig:running_time}
\end{figure}

%% file: tables/experiment_a_b.tex
\begin{table}
\parbox{.48\linewidth}{
    \centering
    \caption{This table shows our Cityscapes val results using different channel numbers $k$ in our grouping modules.}
    \begin{tabular}{ccc|cc}
    \hline 
    $k$=8      & $k$=16     & $k$=32     & AP            & AP50\%        \\ \hline\hline
    \checkmark &            &            & 29.8          & 49.7          \\
               & \checkmark &            & 30.2          & 50.2          \\
               &            & \checkmark & \textbf{32.0} & \textbf{52.0} \\ \hline
    \end{tabular}
    \label{table:experiment_a}
}
\hfill
\parbox{.48\linewidth}{
    \centering
    \caption{This table shows our Cityscapes val results using different combinations of affinity loss weights.}
    \resizebox{0.46\textwidth}{!}{
        \begin{tabular}{cccc|cc}
        \hline 
        $\lambda_{bdr/sld}^{(1)}$ & $\lambda_{bdr/sld}^{(2)}$ & $\lambda_{bdr/sld}^{(3)}$ & $\lambda_{bdr/sld}^{(4)}$ & AP            & AP50\%        \\ \hline\hline
        0.125                     & 0.250                     & 0.500                     & 1.000                        & 30.9          & 51.8          \\
        0.250                     & 0.500                     & 1.000                     & 1.000                     & \textbf{32.0} & \textbf{52.0} \\
        0.500                     & 1.000                     & 1.000                     & 1.000                     & 29.4          & 49.2          \\
        1.000                     & 1.000                     & 1.000                     & 1.000                     & 29.6          & 49.6          \\ \hline
        \end{tabular}
    }
    \label{table:experiment_b}
}
\end{table}

%% file: tables/experiment_c.tex
\begin{table}[t!]
\centering
\caption{This table shows our Cityscapes val results on various modifications of the Cascade-GAEC algorithm. }
\resizebox{0.5\textwidth}{!}{
    \begin{tabular}{c|ccc|cc}
    \hline 
    Affinity   & \multicolumn{3}{c|}{Grouping}        &                               \\ \hline 
    GAEC       & GAEC       & PA-GAEC    & GAS      & AP            & AP50\%        \\ \hline\hline
    \checkmark &            &            &            & 26.6          & 46.5          \\ 
    \checkmark & \checkmark &            &            & 30.4          & 49.2          \\
    \checkmark &            & \checkmark & \checkmark & 32.0          & 52.0          \\
    \checkmark &            & \checkmark &            & \textbf{32.4} & \textbf{53.1} \\ \hline
    \end{tabular}
}
\vspace{-0.3cm}
\label{table:experiment_c}
\end{table}

%% file: TEX/5_discussion.tex
\label{sec:discussion}

There are two questions that need to be further discussed. One is how to design robust networks and algorithms to parse oversized and undersized instances. The other is how to increase networks' capability with more advanced affinities. 

\textbf{Extreme-size instances} are hard cases for both region-based methods and proposal-free methods. Region-based approaches attempt to solve these cases by adding more anchors that vary in size and output more RoIs, but its effectiveness on extreme-size instances is limited and its computational cost is high. While this challenge has partially solved by single-stage approaches with feature pyramids, its prediction power on extreme-size instances is limited by the pyramid level it uses. Therefore, we need more ideas on how to solve this challenge neatly. 

\textbf{Advanced affinities} such as coexistence and co-location between instances haven't been fully explored through nowadays literature. In DaffNet, we demonstrate the effectiveness of combining pixel affinities with fixed grid spacing and segment affinities with non-fixed grid spacing. Yet, more research is needed on developing affinity-based models that understand the context of the image in a more intellectual way.

%% file: TEX/6_conclusion.tex
In this work, we introduce a new affinity-based instance segmentation network: Deep Affinity Net (DaffNet) and we propose a new graph partitioning algorithm: Cascade-GAEC. The single-shot performance of DaffNet reaches \resultval AP on Cityscapes val and \resulttest AP on test, surpassing all previous affinity-based networks as well as the milestone region-based model Mask R-CNN. In conclusion, we prove that, without using regions or keypoints, instance segmentation tasks can also be solved effectively and efficiently via affinity.

%% file: TEX/appendix.tex
\section*{Appendices}
\renewcommand{\thesubsection}{\Alph{subsection}}

\subsection{Greedy Association}
\label{appex:1}

Recall that in Sec~\ref{sec:3_5}, we introduce a helper function, namely Greedy Association (GAS), to help accelerating our Cascade-GAEC.  The purpose of GAS is to avoid the expensive $O(\log n)$ priority queue insert operation in GAEC when the input graph contains a large number of vertices. Instead, GAS traverse all unlabeled vertices in random order and assign them with the labels from their closest associated neighbors (i.e. the neighbors with the largest affinities). The detail of GAS algorithm is shown below in Alg.~\ref{alg:gas}.

\input{algorithms/gas}

There is a trade off between speed and accuracy on whether involve GAS in Cascade-GAEC or not. When using GAS, our experiments result a 0.4\% AP decrement on Cityscapes val and 0.5\% AP decrement on test. On the other hand, GAS lower the running time for 0.14 second per image on our Cascade-GAEC algorithm.

%% file: algorithms/gas.tex
\begin{algorithm}
    Set $G=(V,E)$\;
    Set $V_n \in V$ unlabeled vertices\;
    \While {$V_n\ne\emptyset$ and $V_n$ changed} {
        \For {$v \in V_n$} {
            $V_v^{adj}$ = all adjacency vertices of $v$\;
            $u = \argmax_u(a_{v,u}), u\in V_v^{adj}, u\notin V_n$\;
            \If {$a_{v,u}<$ threshold} {
                Continue\;
            }
            label $v$ as $u$\;
            $V_n = V_n \setminus v$\;
        }
    }
    label $v\in V_n$ as background\;
    \Return $L_V$\;
    \caption{Greedy Association (GAS)}
    \label{alg:gas}
\end{algorithm} 